\documentclass[10pt,twocolumn,letterpaper]{article}

\usepackage{cvpr}              %

\usepackage{booktabs}  
\usepackage{multirow}  
\usepackage{rotating}  
\usepackage{array}
\usepackage{graphicx}  
\usepackage{siunitx}  
\usepackage{diagbox}
\usepackage{longtable}  
\usepackage{float}
\usepackage{xspace}
\usepackage{xcolor}
\usepackage[accsupp]{axessibility}  %

\newcommand{\ourmethod}{TARDIS\xspace}

\definecolor{cvprblue}{rgb}{0.21,0.49,0.74}
\usepackage[pagebackref,breaklinks,colorlinks,allcolors=cvprblue]{hyperref}

\def\paperID{25} %
\def\confName{EarthVision}
\def\confYear{2025}

\title{Distribution Shifts at Scale: Out-of-distribution Detection in Earth Observation}

\author{
Burak Ekim\thanks{Work done while a resident at the Microsoft AI for Good Research Lab. Contact: {\tt\small burak.ekim@unibw.de}}\\
University of the Bundeswehr Munich
\and
Girmaw Abebe Tadesse\thanks{Residency supervisor.}\\
Microsoft AI for Good Research Lab\\
\and
Caleb Robinson\footnotemark[2]\\
Microsoft AI for Good Research Lab\\
\and
Gilles Hacheme\\
Microsoft AI for Good Research Lab\\
\and
Michael Schmitt\\
University of the Bundeswehr Munich\\
\and
Rahul Dodhia\\
Microsoft AI for Good Research Lab\\
\and
Juan M. Lavista Ferres\\
Microsoft AI for Good Research Lab\\
}

\begin{document}
\maketitle
\begin{abstract}
Training robust deep learning models is crucial in Earth Observation, where globally deployed models often face distribution shifts that degrade performance, especially in low-data regions. Out-of-distribution (OOD) detection addresses this by identifying inputs that deviate from in-distribution (ID) data. However, existing methods either assume access to OOD data or compromise primary task performance, limiting real-world use. We introduce TARDIS, a post-hoc OOD detection method designed for scalable geospatial deployment. Our core innovation lies in generating surrogate distribution labels by leveraging ID data within the feature space. TARDIS takes a pre-trained model, ID data, and data from an unknown distribution (WILD), separates WILD into surrogate ID and OOD labels based on internal activations, and trains a binary classifier to detect distribution shifts. We validate on EuroSAT and xBD across 17 setups covering covariate and semantic shifts, showing near-upper-bound surrogate labeling performance in 13 cases and matching the performance of top post-hoc activation- and scoring-based methods. Finally, deploying TARDIS on Fields of the World reveals actionable insights into pre-trained model behavior at scale. The code is available at \href{https://github.com/microsoft/geospatial-ood-detection}{https://github.com/microsoft/geospatial-ood-detection}
\end{abstract}

\begin{figure}[h]
    \centering
    \includegraphics[width=0.92\linewidth]{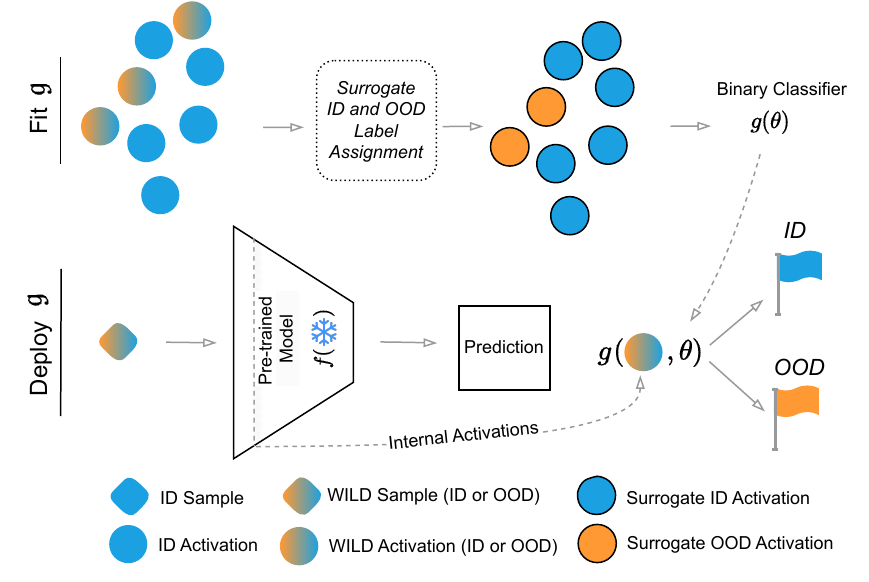}
    \caption{Overview of the proposed OOD detection method. Given a pre-trained model, ID samples, and WILD samples (from unknown distributions), TARDIS assigns surrogate ID/OOD labels to WILD samples using the ID set and fits a binary classifier \( g \) (top row). During deployment, \( g \) uses internal activations of unseen samples to predict whether they are ID or OOD (bottom row).}
    \label{fig:overview_illustration}
\end{figure}

\section{Introduction}

Deep learning models have demonstrated remarkable capabilities across various domains but often exhibit overconfidence in their predictions, even when confronted with data that diverges from their training distribution~\cite{overconfident_szegedy2013intriguing, overconfident_hendrycks2016baseline, overconfident_nguyen2015deep}. This overconfidence arises from the assumption that inference data will follow the same independent and identically distributed (\textit{i.i.d.}) properties as the training data. However, in real-world applications, this closed-world assumption~\cite{closed_world_he_15_iccv, closed_world_NIPS2012_c399862d} is frequently violated by test-time distribution shifts (e.g., lighting or angle variations, different devices or sensors) that can significantly degrade model performance and harm generalization. To address this, it is critical for predictive models to detect when new observations fall outside the training distribution, a task known as out-of-distribution (OOD) detection. Differentiating OOD from in-distribution (ID) samples is particularly challenging, partly due to the poor calibration of neural networks~\cite{calibration_guo2017, calibration_minderer2021revisitingcalibrationmodernneural}. This degradation poses a critical challenge, especially in applications where incorrect predictions can have severe consequences.

In Earth Observation (EO), specifically with satellite imagery, distribution shifts between training and inference commonly occur in the form of covariate and semantic shifts. Covariate shift refers to changes in the input distribution while keeping the task or label space fixed, whereas semantic shift involves changes in the label distribution or the meaning of inputs. A specific case of covariate shift is known as subpopulation shift~\cite{koh2021wilds}, where the overall data distribution changes due to different proportions of underlying subgroups, even though the subgroups themselves remain the same. In EO, this arises when certain seasons, regions, or acquisition conditions are underrepresented during training but appear more prominently at test time. The magnitude of distribution shifts can further be characterized along a spectrum from near-distribution to far-distribution shifts. Near-distribution shifts may occur within the same satellite source due to temporal, geographic, or environmental variation—such as seasonal changes, sensor degradation, or atmospheric effects like cloud cover~\cite{poortransfer_kerner2024multi}. Far-distribution shifts involve more substantial differences, such as changes in satellite platforms, sensor types, or entirely different imagery domains, resulting in greater divergence from the training distribution. This variability necessitates tailored approaches for satellite-based machine learning models~\cite{mission_critical_rolf2024}.

Despite the vital role of robust models in EO, few studies explore OOD detection in this domain~\cite{eo_ood_zhu_im_cslf, eo_ood_datcu_burned_area, eo_ood_audebert_diffusion, xie2020n, Gawlikowski2023}. These works often presume access to test-time distributions, adhere to closed-set assumptions, and/or degrade model performance on in-distribution tasks. This underscores the need for specialized OOD detection methods suited for large-scale EO deployment. A further challenge arises when applying geospatial models globally. Global geospatial models often suffer significant performance drops in low-data regions \cite{aiken2023fairness}, such as sub-Saharan Africa~\cite{subsaharan_kerner2024accurate}. These issues emphasize the need for global models equipped to detect OOD samples during inference.

\begin{figure*}[h!]
    \centering
    \includegraphics[width=1.0\linewidth]{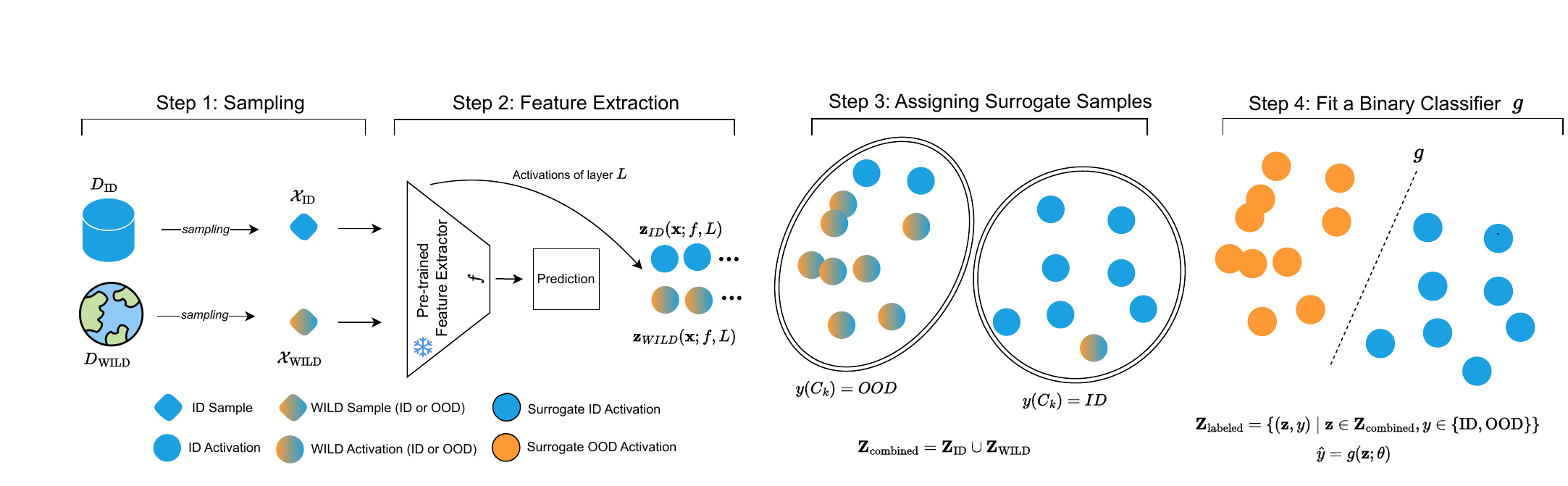}
    \caption{The proposed framework consists of four key steps: (1) Sampling in-distribution (ID) and WILD samples; (2) Extracting internal activations from a pre-trained model  \( f \) for both ID and WILD samples; (3) Clustering the combined feature space and labeling WILD samples as surrogate-ID or surrogate-OOD; (4) Fitting a binary classifier \( g \) on the labeled feature representations to distinguish between ID and OOD samples. The classifier \( g \), during deployment, flags out-of-distribution inputs.}
    \label{fig:full_flow}
\end{figure*}

In this paper, we build on well-established finding that OOD samples trigger internal activation patterns that diverge from ID samples \cite{react_sun2021, nap_ood}. Building on this insight and recognizing the challenges of deploying geospatial models at scale, we introduce the following key contributions:
\begin{itemize}
\item We propose \ourmethod (Test-time Addressing of Distribution Shifts at Scale), a lightweight post-hoc OOD detection method that preserves model performance and operates without access to labeled data from unknown test-time shifts.
\item We evaluate \ourmethod on EuroSAT (patch classification) and xBD (semantic segmentation) under near-distribution settings, analyzing covariate (geographical, temporal, environmental) and semantic (withheld class) shifts across 17 experimental setups. We also compare its performance against existing post-hoc OOD detection methods.
\item We show that \ourmethod scales effectively, providing actionable insights into the robustness and trustworthiness of geospatial models in real-world scenarios.  
\end{itemize}

\section{Related Works}
\label{sec:relatedworks}

Our method lies at the intersection of two notable categories of OOD detection methods: scoring functions and activation manipulation. Scoring functions assign a numerical score to each input, reflecting its alignment with ID samples based on the model's output. Maximum Softmax Probability (MSP)~\cite{overconfident_hendrycks2016baseline} detects OOD samples by assessing the softmax confidence score, assuming that low confidence indicates OOD samples. The Out-of-DIstribution detector for Neural networks (ODIN)~\cite{odin_liang2020} enhances MSP by applying input perturbations and temperature scaling to improve ID-OOD separation. The Mahalanobis score~\cite{mahalo_lee2018} calculates the distance between input features and class means in feature space, flagging inputs far from these means as OOD. The energy score~\cite{energy_liu2021} evaluates the model's energy function to assess the likelihood of an input belonging to the ID distribution. Another category of OOD detection methods manipulates the internal activations of a pre-trained model to improve detection performance. ReAct~\cite{react_sun2021} identifies differences in activation patterns of the penultimate layer between ID and OOD samples, enhancing separation by clipping activations at an upper limit. DICE~\cite{dice_sun2022} applies weight sparsification on a specific layer to further distinguish ID from OOD data and, when combined with ReAct, can enhance detection performance. Activation Shaping~\cite{ash_djurisic23} prunes a portion of an input sample's activations and slightly adjusts the remaining activations; when paired with the energy score, this approach has been shown to outperform contemporary OOD methods. Neuron Activation Patterns (NAP)~\cite{nap_ood} extracts, downsamples, and binarizes activation patterns from convolutional layers, computing the smallest Hamming distance between binarized test and training patterns. This distance provides a measure of model uncertainty, aiding in OOD detection. Both scoring functions and activation-based strategies face notable challenges in dynamic, real-world environments. Scoring functions, while maintaining ID accuracy, struggle with OOD detection due to their reliance on static data distributions, an assumption rarely valid in practice. Activation-based methods, though effective, often compromise ID task performance and rely on hyperparameters tuned with hypothetical or substitute OOD samples. This reliance limits their effectiveness under evolving test-time conditions and shifting distributions.

Another line of OOD detection research uses clustering in feature space to separate ID and OOD samples. These methods exploit structural patterns in learned representations rather than direct scoring or activation manipulation. Gulati et al.~\cite{gulati2024out} improve over \( k \)-means by using non-negative kernel regression for soft clustering. Sinhamahapatra et al.~\cite{sinhamahapatra2022all} explore clustering in self-supervised latent spaces, showing its potential in low-label settings. Closely related in spirit, NG-Mix~\cite{dong2024nng} proposes pseudo-anomaly generation for semi-supervised anomaly detection. 

The task of OOD detection is inherently challenging, and the EO domain is no exception due to the diverse and heterogeneous nature of satellite imagery. Despite the critical importance of OOD detection in EO, this area remains relatively underexplored, with only a limited number of studies addressing the issue. One such study develops a Dirichlet Prior Network to quantify distributional uncertainty in deep learning models for satellite image analysis~\cite{eo_ood_zhu_im_cslf}. This method assumes Dirichlet distributions for ID samples and employs various setups where classes, color channels, and environmental features are alternately treated as ID or OOD. Another study frames anomaly detection as a cumulative open-set detection and location separation task~\cite{eo_ood_datcu_burned_area}. This approach also uses a Dirichlet prior, like the first study, to expand the representation space between ID (normal) and OOD (anomalous) samples by predicting the anticipated categorical distribution. The method suits scenarios where pre-event images are unavailable, subject to radiation differences, or not recent enough to aid detection.

In this paper, we do not aim to optimize OOD detection performance but propose a method suited for real-world applications where the distribution is unknown, maintaining ID task performance is critical, and expensive methods are impractical. Our method detects shifts within the same satellite sources. This type of near-distribution shift presents a greater challenge than far-distribution shifts, such as those between natural and satellite imagery, as the differences are more subtle and harder to disentangle. By tackling these nuanced shifts, our method makes OOD detection more practical for real-world geospatial applications.

\section{Problem Formulation}
\label{subsec:problem_formulation}

Let \( \mathcal{X} \) represent the set of all possible data the model may encounter, and \( \mathcal{Y} \) the set of class labels. The dataset on which a model is trained is defined as in-distribution (ID), denoted \( D_{\text{ID}} \subset \mathcal{X}_{\text{ID}} \times \mathcal{Y} \). During inference, however, the model may encounter data from an unknown distribution, referred to as the WILD dataset \( D_{\text{WILD}} \subset \mathcal{X}_{\text{WILD}} \), which may contain both ID and OOD samples.

We assume a pre-trained neural network \( f: \mathcal{X} \rightarrow \mathbb{R}^{|\mathcal{Y}|} \), trained on \( D_{\text{ID}} \), and our objective is to distinguish between ID and OOD samples within \( D_{\text{WILD}} \). The network \( f \) extracts features \( \mathbf{z} \in \mathbb{R}^F \), where \( F \) is the dimensionality of the feature space. Specifically, we fit a binary classifier \( g: \mathbb{R}^F \rightarrow \{0, 1\} \), parametrized by \( \theta \), that operates on these features \( \mathbf{z} \) for a given sample in \( \mathcal{X}_{\text{WILD}} \). We define \( g(\mathbf{z}; \theta) = 0 \) if \( \mathbf{x} \sim D_{\text{ID}} \), and \( g(\mathbf{z}; \theta) = 1 \) if \( \mathbf{x} \sim D_{\text{OOD}} \).

\begin{table*}[ht]
    \centering  
    \caption{Overview of experimental setups for evaluating \ourmethod under controlled distribution shifts. The table details the covariate and semantic shift scenarios, including the dataset used, split method, and the composition of the ID and out-of-distribution OOD sets. All conditions are post-event except where specified.}  
    \label{tab:covariate_shifts}  
    \begin{tabular}{>{\centering\arraybackslash}m{0.2cm} >{\centering\arraybackslash}m{0.2cm} m{5.5cm} m{5cm} m{5cm}}  
        \toprule  
        \textbf{\begin{sideways}Type\end{sideways}} & \textbf{\begin{sideways}Dataset\end{sideways}} & \textbf{Split Method} & \textbf{ID (Training Set)} & \textbf{OOD (Test Set)} \\  
        \midrule  
        \multirow{6}{*}{\begin{sideways}Covariate Shift\end{sideways}} & \begin{sideways}EuroSAT\end{sideways} & Spatial Split (Longitude) & West Europe & East Europe \\  
        \cmidrule{2-5}  
        & \multirow{5}{*}{\begin{sideways}xBD\end{sideways}} & Similar Disasters - Distant Locations & Nepal Flooding & Midwest Flooding \\  
        \cmidrule{3-5}  
        & & Similar Disasters - Nearby Locations & Santa Rosa Wildfire & Woolsey Fire \\  
        \cmidrule{3-5}  
        & & Different Disasters - Distant Locations & Hurricane Matthew & Nepal Flooding \\  
        \cmidrule{3-5}  
        & & Different Disasters - Nearby Locations & Hurricane Matthew & Mexico Earthquake \\  
        \cmidrule{3-5}  
        & & Temporal & Portugal Wildfire (Pre-Disaster) & Portugal Wildfire \\  
        \midrule[\heavyrulewidth]  
        \begin{sideways}Semantic Shift\end{sideways} & \begin{sideways}EuroSAT\end{sideways} & Unseen Class \newline (Repeated for 10 Classes) & 9 out of 10 classes (e.g., Annual Crop, Forest, Herbaceous Vegetation, Highway, Industrial, Pasture, Permanent Crop, Residential, River) & 1 holdout class (e.g., Sea/Lake) \\  
        \bottomrule  
    \end{tabular}  
\end{table*}  

\section{Method}
\label{sec:method}  

Training a classifier to distinguish in-distribution (ID) from out-of-distribution (OOD) samples typically requires labeled ID and OOD data, which is impractical in real-world settings. This motivates generating surrogate labels for samples with unknown distributions, referred to as WILD samples. Our method, illustrated in Figure~\ref{fig:overview_illustration}, operates with a pre-trained model, known ID samples, and WILD samples. Using known ID samples, it assigns surrogate ID and OOD labels to WILD samples, which are then used to train a binary distribution classifier. \ourmethod involves training a binary classifier to discriminate between ID and OOD samples based on internal activations, as shown in Figure~\ref{fig:full_flow}. Given \( M \) ID and \( N \) WILD samples, we pass both through the frozen model and extract activations from a specified layer. After pooling for spatial downsampling, we obtain one-dimensional feature vectors, combine them, and assign corresponding ID or WILD labels. We shuffle this set and reserve 30\% for validation. On the remaining 70\%, we run \( k \)-means clustering to assign surrogate ID/OOD labels to WILD samples based on distance to ID samples. Using this labeled set, we train a binary classifier \( g(\mathbf{z}; \theta) \) to map activations to distribution labels.

At deployment, we extract activations from a sample of unknown distribution and predict its label as \( \hat{y} = g\left( \text{Downsample}\left( f(\mathbf{x}) \right); \theta \right) \). The output can be interpreted as a probability of domain shift or thresholded for binary classification (\( 0 \): ID, \( 1 \): OOD).

\textbf{Surrogate Label Assignment.} The core novelty of \ourmethod lies in generating surrogate labels for WILD samples based on their proximity to known ID samples in feature space. We assume that ID and OOD features are separable enough for clustering to be effective. Specifically, we cluster the combined ID+WILD feature space into \( k \) groups (Step 3 in Figure~\ref{fig:full_flow}). For each cluster, if the proportion of ID samples is greater than or equal to the threshold \( T \), all samples in that cluster receive label 0 (ID); otherwise, label 1 (OOD). Distance to ID samples thus serves as a proxy for assigning surrogate labels.

Given \( k \), \( T \), and the feature vectors, we perform clustering and assign surrogate labels accordingly. To select the optimal parameters, we minimize the composite objective:

\begin{equation*}
(k^*, T^*) = \arg\min_{k, T} \quad H(S) + P_{\text{mis-ID}} - P_{\text{corr-ID}}
\end{equation*}

where \( H(S) \) is the average cluster entropy (encouraging homogeneity), \( P_{\text{mis-ID}} \) is the proportion of ID samples misclassified as OOD, and \( P_{\text{corr-ID}} \) is the proportion of correctly classified ID samples. This balances three goals: coherent clusters, low ID misclassification, and high correct ID classification, promoting better surrogate label quality.

\begin{table*}[ht]  
    \centering  
    \caption{Performance metrics comparing the oracle classifier \( g_{oracle} \) and the surrogate classifier \( g^* \) for EuroSAT and xBD datasets across various experimental setups. The \( g_{oracle} \) classifier acts as an upper bound for the \( g^* \), which is the proposed method. The † notation indicates that over 10 measurements, the difference was not found to be statistically significant (\( p < 0.05 \)). All conditions are post-event except where specified.}  
    \begin{tabular}{>{\centering\arraybackslash}m{0.1cm} >{\centering\arraybackslash}m{0.1cm} p{3.5cm} p{3.2cm} c c c c}  
        \toprule  
        \textbf{\begin{sideways}Type\end{sideways}} & \textbf{\begin{sideways}Dataset\end{sideways}} & \textbf{OOD (Test Set)} & \textbf{ID (Training Set)} & \multicolumn{2}{c}{\textbf{AUROC$\uparrow$}} & \multicolumn{2}{c}{\textbf{FPR95$\downarrow$}} \\  
        & & & & \textbf{\( g_{oracle} \)} & \textbf{\( g^* \)} & \textbf{\( g_{oracle} \)} & \textbf{\( g^* \)} \\  
        \midrule  
        \multirow{7}{*}{\begin{sideways}Covariate Shift\end{sideways}} & \begin{sideways}EuroSAT\end{sideways} & Eastern Europe & Western Europe & $0.91 \pm 0.04$† & $0.89 \pm 0.06$† & $0.32 \pm 0.11$† & $0.37 \pm 0.10$† \\  
        \cmidrule{2-8}  
        & \multirow{6}{*}{\begin{sideways}xBD\end{sideways}} & Midwest Flooding & Nepal Flooding & $1.00 \pm 0.00$† & $1.00 \pm 0.00$† & $0.01 \pm 0.02$† & $0.01 \pm 0.02$† \\  
        & & Woolsey Fire & Santa Rosa Wildfire & $0.94 \pm 0.02$† & $0.93 \pm 0.02$† & $0.23 \pm 0.08$† & $0.27 \pm 0.09$† \\  
        & & Nepal Flooding & Hurricane Matthew & $0.99 \pm 0.01$ & $0.98 \pm 0.01$ & $0.06 \pm 0.04$† & $0.07 \pm 0.02$† \\  
        & & Mexico Earthquake & Hurricane Matthew & $0.96 \pm 0.02$† & $0.94 \pm 0.03$† & $0.11 \pm 0.05$† & $0.19 \pm 0.08$† \\  
        & & Portugal Wildfire (Pre) & Portugal Wildfire & $0.99 \pm 0.00$ & $0.95 \pm 0.01$ & $0.06 \pm 0.04$ & $0.23 \pm 0.20$ \\  
        & & Santa Rosa Wildfire & Woolsey Fire & $0.94 \pm 0.02$† & $0.93 \pm 0.02$† & $0.23 \pm 0.08$† & $0.27 \pm 0.09$† \\  
        \midrule[\heavyrulewidth]  
        \multirow{10}{*}{\begin{sideways}Semantic Shift\end{sideways}} & \multirow{10}{*}{\begin{sideways}EuroSAT\end{sideways}} & Forest & \multirow{10}{*}{\rotatebox[origin=c]{90}{Remaining 9 classes}} & $0.99 \pm 0.01$† & $0.97 \pm 0.02$† & $0.08 \pm 0.04$† & $0.08 \pm 0.05$† \\  
        & & Herbaceous Vegetation & & $0.93 \pm 0.03$† & $0.88 \pm 0.07$† & $0.26 \pm 0.11$† & $0.29 \pm 0.14$† \\  
        & & Highway & & $0.63 \pm 0.08$† & $0.56 \pm 0.06$† & $0.78 \pm 0.13$† & $0.88 \pm 0.10$† \\  
        & & Industrial & & $0.97 \pm 0.02$† & $0.96 \pm 0.02$† & $0.13 \pm 0.14$† & $0.22 \pm 0.07$† \\  
        & & Pasture & & $0.98 \pm 0.01$ & $0.95 \pm 0.01$ & $0.09 \pm 0.06$† & $0.28 \pm 0.19$† \\  
        & & Permanent Crop & & $0.92 \pm 0.03$† & $0.89 \pm 0.03$† & $0.26 \pm 0.08$† & $0.33 \pm 0.11$† \\  
        & & Residential & & $0.87 \pm 0.03$† & $0.82 \pm 0.06$† & $0.46 \pm 0.21$† & $0.43 \pm 0.15$† \\  
        & & River & & $0.87 \pm 0.04$ & $0.81 \pm 0.06$ & $0.44 \pm 0.14$† & $0.57 \pm 0.10$† \\  
        & & Sea-Lake & & $1.00 \pm 0.00$† & $1.00 \pm 0.00$† & $0.00 \pm 0.00$† & $0.00 \pm 0.00$† \\  
        & & Annual Crop & & $0.92 \pm 0.03$† & $0.91 \pm 0.04$† & $0.25 \pm 0.10$† & $0.32 \pm 0.16$† \\  
        \bottomrule  
    \end{tabular}  
    \label{tab:exp_results_combined}  
\end{table*}  

\section{Experimental Setup}
\label{section:expsetup}

In this section, we describe experimental setups to evaluate \ourmethod under controlled distribution shifts, mimicking real-world geospatial deployment challenges. We use the EuroSAT~\cite{eurosat} dataset for patch-level classification and the xBD~\cite{xBD} dataset for semantic segmentation. EuroSAT contains 27,000 labeled images across ten land use/land cover classes. xBD includes pre/post-disaster imagery labeled for building damage assessment, covering hurricanes, floods, wildfires, and more. For our purposes, we formulate xBD as a building footprint segmentation task with two classes: building footprint and background. These datasets offer diverse evaluation conditions in terms of volume, task, sensor, spatial/temporal coverage, and geolocation, making them suitable test-beds for our OOD method.

To simulate controlled distribution shifts, we adjust the train, validation, and test splits in both datasets so that the model, trained on the modified training set, encounters realistic shifts during testing. We evaluate two shift types: covariate shift, where input distribution changes between train and test, and semantic shift, where unseen classes appear at test time. Covariate shifts include spatial (different locations), temporal (different time periods), and environmental (different landcover/disaster types) variance. For semantic shift, we train on nine of ten classes and test on the held-out class, rotating through all classes. Table~\ref{tab:covariate_shifts} summarizes these setups.

These setups assume access to clear ID (train) and OOD (test) labels—an unrealistic assumption in real-world deployments. To address this, we evaluate our method in two ways. First, we use known ID and OOD labels to directly train and evaluate the binary classifier \( g \), referred to as \( g_{\text{oracle}} \). Second, we treat test-time OOD labels as unknown (i.e., WILD) and apply our surrogate label assignment to generate surrogate ID and OOD labels. This version, trained and evaluated using surrogate labels, is denoted as \( g^* \). We consider \( g_{\text{oracle}} \) as an upper bound on performance, since it uses ground-truth labels, while \( g^* \) relies on inferred surrogates.

\section{Experimental Results}
\label{section:expsresults}

In this section, we present ablation studies and experimental results. Note that ID task performance is not reported, as it remains unaffected by our method.

\textbf{Ablation Studies}. We conduct benchmark studies on the \( g_{\text{oracle}} \) classifier to explore the factors influencing \ourmethod's ability to detect test-time distribution shifts. This ability depends significantly on the layer from which internal activations are extracted and the downsampling method applied. To address this, we test different layers and compare their performance (Table \ref{suptab:layer_benchmark_eurosat} and Table \ref{suptab:layer_benchmark_xbd}). For downsampling, we experiment with several methods, including mean and standard deviations, average pooling, max pooling, and PCA-based reduction. The max pooling-based downsampling method achieves the highest performance, likely due to its ability to retain the most salient activation patterns, which we argue is important for effective OOD detection (Table \ref{suptab:downsampling_benchmark}). We then evaluate various classifiers: KNeighbors, GaussianNB, DecisionTree, ExtraTrees, LogisticRegression, SVC, RandomForestUnbalanced, RandomForest, AdaBoost, and GradientBoosting. Results indicate that Logistic Regression provides the best tradeoff between classification performance and wall time (Table \ref{suptab:g_benchmark}). We then compare \( g^* \) to \( g_{\text{oracle}} \) as we tune the parameters \(k\) and \( T \) required for surrogate label assignment. To determine \(k\) and \( T \), we use a Tree-structured Parzen Estimator for sampling. Specifically, we set the search boundaries for \(k\) (the number of clusters) between 2 and 0.3 $\times$ \( M \), where \( M \) is the total number of samples, and for \( T \) (the fraction of in-distribution samples within a cluster, below which the cluster is labeled as surrogate OOD) between 0.01 and 0.2. These values were empirically determined based on preliminary tuning for optimal performance. We then run 20 independent experiments and select the best-performing \( (k, T) \) pair. Fixing \( T \) based on the hyperparameter search results, we observe a recurring pattern that enables us to fix \(k\) to 0.3 $\times$ M across all setups. This choice is based on the observation that the classifier \( g^* \) reaches performance levels close to \( g_{\text{oracle}} \) when \( k \) is set to 30\% of the training samples. This trend holds for both datasets, with the xBD disaster scenario (semantic shift) and the EuroSAT Pasture experiment (covariate shift) showing similar behavior. In both cases, performance improves as more clusters are added, approaching that of \( g_{\text{oracle}} \). In Figures~\ref{supfig:optuna_eurosat} and~\ref{supfig:optuna_xbd}, we show how we determine the optimal \( (k, T) \) pair that achieves high surrogate label assignment performance across all setups. The same pair generalizes to real-world deployment, removing the need for hyperparameter tuning.

\textbf{Surrogate Sample Assignment Benchmark.} Table~\ref{tab:exp_results_combined} shows results for \( g^* \) and \( g_{\text{oracle}} \) across 17 setups. Overall, \( g^* \) closely approaches \( g_{\text{oracle}} \), which serves as the upper bound. In 13 setups for AUROC and all 17 for FPR95, the performance gap is not statistically significant. In the few remaining cases, the gap is small. These results show that \ourmethod can effectively assign surrogate labels without access to test-time labels.

\textbf{ID/OOD Balance.}  
As discussed in Section~\ref{sec:method}, splitting ID and WILD data into validation and clustering sets affects the ID/OOD ratio throughout the pipeline. In Table~\ref{tab:exp_results_combined}, WILD consists entirely of OOD samples, but in real deployments, it may contain both ID and OOD. While data is shuffled, monitoring this ratio is key to understanding sensitivity. For example, in the \textit{Pasture} semantic shift setup, the clustering and classifier stages see 17\% and 15\% OOD, respectively. In \textit{Nepal Flood – Midwest Flood}, the ratios are 46\% and 35\%. This variation shows that TARDIS performs robustly across different ID/OOD balances, supporting the generalizability of our method.

\begin{figure*}[ht]
    \centering
    \includegraphics[width=0.90\linewidth]{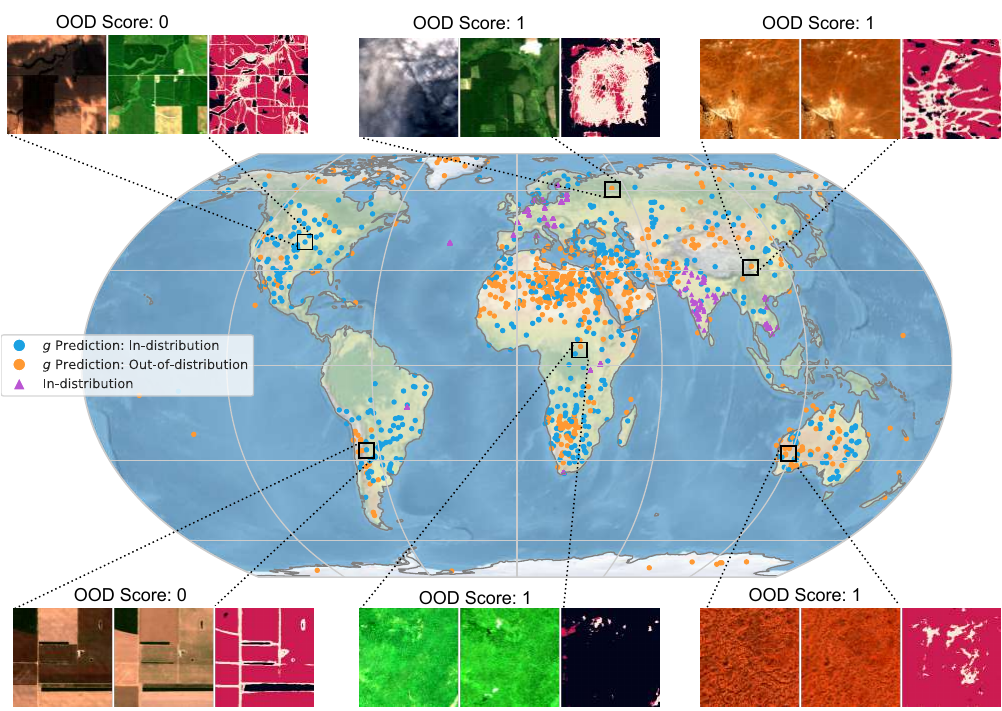}
    \caption{Geographical distribution of ID and WILD sets, containing 500 and 1200 samples, respectively. The ID set is sampled from the FTW dataset training set, while the WILD set is randomly sampled from the Microsoft Planetary Computer. Each Sentinel-2 patch, provided in two different time frames (planting and harvesting). The model \( f \) takes both Sentinel-2 images from different seasons as input, and the predictions are shown on the right. The OOD probability values are thresholded at 0.5.}
    \label{fig:deployment_distribution_map}
\end{figure*}

\begin{table}[ht]  
    \centering  
    \caption{Performance comparison of post-hoc score-based and activation-based OOD detection methods across 10 semantic shift and 7 covariate shift experiments on EuroSAT and xBD. We report the mean ± standard deviation across all 10 or 7 experiments for each method. \ourmethod achieves comparable performance while requiring no hyperparameter tuning and preserving main task performance.}
    \label{tab:benchmarking_combined}  
    \begin{tabular}{>{\centering\arraybackslash}m{0.1cm} >{\centering\arraybackslash}m{3cm} >{\centering\arraybackslash}m{1.9cm} >{\centering\arraybackslash}m{1.8cm}}
        \toprule  
        \textbf{\begin{sideways}\end{sideways}} & \textbf{Method} & \textbf{AUROC$\uparrow$} & \textbf{FPR95$\downarrow$} \\  
        \midrule  
        \multirow{6}{*}{\begin{sideways}Semantic Shift\end{sideways}} 
        & Mahalanobis \cite{mahalo_lee2018} & 0.85 $\pm$ 0.01 & 0.19 $\pm$ 0.05  \\  
        & MSP \cite{overconfident_hendrycks2016baseline} & 0.90 $\pm$ 0.03 & 0.23 $\pm$ 0.07 \\  
        & Energy \cite{energy} & 0.89 $\pm$ 0.02 & 0.20 $\pm$ 0.04  \\ 
        & ReAct \cite{react_sun2021} & 0.92 $\pm$ 0.04 & 0.13 $\pm$ 0.01 \\  
        & NAP \cite{nap_ood} & 0.93 $\pm$ 0.03 & 0.21 $\pm$ 0.05 \\  
        & TARDIS (Ours) & 0.95 $\pm$ 0.02 & 0.16 $\pm$ 0.03  \\  
        \midrule  
        \multirow{7}{*}{\begin{sideways}Covariate Shift\end{sideways}} 
        & Mahalanobis \cite{mahalo_lee2018} & 0.88 $\pm$ 0.03 & 0.22 $\pm$ 0.06  \\  
        & MSP \cite{overconfident_hendrycks2016baseline} & 0.92 $\pm$ 0.04 & 0.25 $\pm$ 0.08 \\  
        & Energy \cite{energy} & 0.91 $\pm$ 0.03 & 0.22 $\pm$ 0.05  \\ 
        & ReAct \cite{react_sun2021} & 0.94 $\pm$ 0.05 & 0.15 $\pm$ 0.02 \\  
        & NAP \cite{nap_ood} & 0.95 $\pm$ 0.02 & 0.23 $\pm$ 0.06 \\  
        & TARDIS (Ours) & 0.98 $\pm$ 0.01 & 0.11 $\pm$ 0.04  \\ 
        \bottomrule  
    \end{tabular}  
\end{table}  

\textbf{Comparison with Existing OOD Detection Methods.}
We benchmark existing post-hoc activation-based and score-based methods across all semantic and covariate shift experiments, reporting mean and standard deviation (Table~\ref{tab:benchmarking_combined}). Gradient-based methods involving backpropagation are excluded due to their high computational cost and increased wall time. Score-based methods (Mahalanobis, MSP, Energy) preserve ID performance as they require no model changes, but rely on potentially overconfident predictions and are less sensitive to near-domain shifts. Activation-based methods like ReAct and NAP modify internal activations—via clipping or purification—boosting OOD detection but at the cost of reduced ID performance, limiting their practicality.

\textbf{Why Clustering Alone is Insufficient.}
Our method includes a distance-based clustering step followed by an additional classification step (Step 3 and 4 in Figure \ref{fig:full_flow}). Removing the classification step results in a drop in ROC AUC for the \textit{River} experiment from 97\% to 79\%—a trend consistently observed across all experiments. This demonstrates that clustering alone (i.e., distance-based assessment) is insufficient for effective distribution detection, justifying the need for Step 4 in Figure \ref{fig:full_flow}.

\section{OOD Detection Goes Global: Real-World Deployment}
\label{section:oodatscale}

We use the Fields of the World (FTW) ~\cite{ftw} training set and pre-trained models to demonstrate the capabilities of \ourmethod in a large-scale deployment scenario. FTW is a geographically diverse dataset designed for agricultural field segmentation, covering 24 regions across Europe, Africa, Asia, and South America. The dataset contains approximately 70,000 samples, each consisting of multi-date, multi-spectral Sentinel-2 satellite patches paired with three-class semantic segmentation masks (field, field boundary, and background). The task involves segmenting these classes using a pair of Sentinel-2 images --- one for the planting season and one for the harvesting season --- as input. Field boundary data is crucial for global agricultural monitoring, however training large scale models to segment field boundaries from satellite imagery  presents significant challenges due to the geographic diversity of fields, varying crop cycles, and agro-climatic conditions, all of which introduce substantial distribution shifts. This complexity, along with corresponding distribution shifts found in real world imagery, makes the FTW dataset ideal for testing \ourmethod's ability to diagnose model performance during inference. Additionally, the multi-date nature of the dataset is particularly suitable for evaluating models that must handle spatiotemporal variations in satellite imagery.

\textbf{Sampling ID Set.} To form the ID set, we sample 50 patches from each country represented in the training set, ensuring a geographically diverse ID dataset that closely matches the data on which the model was originally trained.

\textbf{Sampling WILD Set.} We collect multispectral Sentinel-2 L2A satellite images using the Microsoft Planetary Computer~\cite{planetarycomputer}. The images are processed to Level-2A (bottom-of-atmosphere) and stored in cloud-optimized GeoTIFF (COG) format. We use four spectral bands: Red (B04), Green (B03), Blue (B02), and Near-Infrared (B08), each with a spatial resolution of 10 meters per pixel. We randomly query 1200 Sentinel-2 scenes from those over land and with cloud coverage $<10\%$. Then, from each scene, we generate random patches of shape ~256 $\times$ 256 $\times$ 4 pixels (filtering out samples with $\geq 10\%$ zero or \textit{NaN} pixels) and query both the planting and harvesting seasons for the same locations. The planting and harvesting dates are fixed based on the hemisphere: April 1 to June 30 and September 1 to November 30 for the Northern Hemisphere, and October 1 to December 31 and March 1 to May 31 for the Southern Hemisphere. The two patches are concatenated along the channel dimension, resulting in a patch of shape 256 $\times$ 256 $\times$ 8 pixels. Following the pre-trained model's convention, the concatenated patch is upsampled to 512 $\times$ 512 $\times$ 8, processed by the model, and then downsampled back to the original dimensions. For surrogate label assignment, we fix the hyperparameters \(k\) and \( T \) to \( 0.3 \times 1200 \) and 0.1, respectively, as suggested by the benchmark study.

The geographical distribution of ID and WILD samples is shown in Figure~\ref{fig:deployment_distribution_map}, with WILD samples further classified into surrogate ID and OOD. A clear pattern emerges: samples from arid biomes—such as the deserts of Inner Australia, the Sahara, and Patagonia—and polar regions, including Icelandic glaciers and the South Pole, are more frequently assigned as OOD. This is likely because the ID samples predominantly represent mesic environments: moderately moist, managed landscapes typical of agricultural areas. In contrast, arid and polar regions exhibit extreme environmental conditions that differ significantly from those in the ID set. This ecological dissimilarity likely drives their classification as surrogate OOD, reflecting the model's sensitivity to environmental context and distribution shifts.

To evaluate scalability, we measure the time required to classify a WILD sample's internal activation as surrogate ID or OOD using the logistic regression classifier \( g \). A \(256 \times 256\) input patch (10\,m resolution, covering \(6.5536\,\text{km}^2\)) takes 0.003 seconds to process. Applied to mainland Africa (\(29.77 \times 10^6\,\text{km}^2\)), the method runs on a single GPU in under 4 hours—a fraction of the time needed for full \( f \)-model inference. This enables spatial diagnostics, identifying where the field boundary model is likely in- or out-of-distribution and where outputs may be unreliable. Such rapid screening is especially useful in low-data regions, where models are typically less robust and OOD detection can guide targeted interventions or data collection.

\begin{figure}[t]
    \centering
    \includegraphics[width=1.0\linewidth]{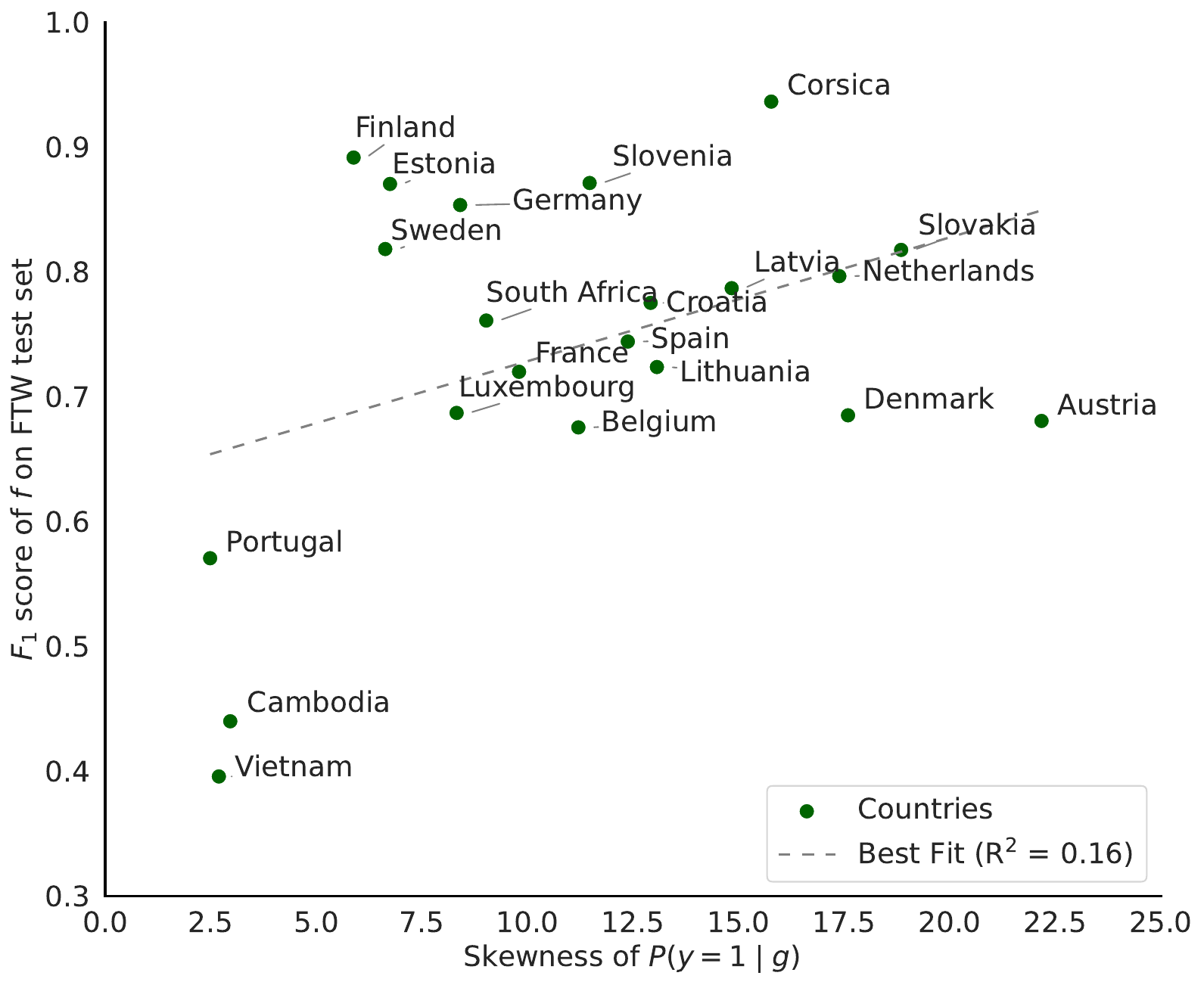}
    \caption{Correlation between model performance and OOD score skewness. Low performance of the \( f \) model on the FTW test set aligns with low skewness in the OOD classifier \( g \)’s score distribution, suggesting the presence of OOD samples in the test set.}
    \label{fig:f1_vs_skew}
\end{figure}

We assess the reliability of the OOD classifier \( g \) by comparing its output skewness on the training set to the performance of the model \( f \) on the FTW test set. High skewness indicates confident (mostly ID or OOD) predictions, while low skewness suggests uncertainty or misclassification. Plotting skewness against \( f \)'s test performance reveals a trend: low skewness aligns with poor model performance (Figure~\ref{fig:f1_vs_skew}), as seen in countries like Portugal, Cambodia, and Vietnam. This suggests that low skewness may signal the presence of OOD samples. Thus, \ourmethod can also serve as a sanity check for detecting distribution shifts in test splits.

\newpage
\textbf{Limitations and Future Work.} Our method assumes access to a non-trivial number of WILD samples and requires periodic updates to the clustering used for surrogate label assignment. Future work could explore density-based clustering methods instead of the current distance-based approach, as they may better handle variable sample densities and outliers, further improving surrogate label assignment.

\section{Conclusion} 
\label{section:conclusion}
We present \ourmethod, a distribution shift detector that assigns surrogate ID and OOD labels to samples from unknown distributions by leveraging clustering in the feature space. These surrogate labels are assigned based on known ID samples and used to train a binary classifier that outputs a distribution score during inference. Rather than optimizing for OOD detection performance, \ourmethod targets real-world scenarios where the distribution is unknown, ID task performance must be preserved, and computational overhead is a concern. Across 13 of 17 experiments mimicking real-world semantic and covariate distribution shifts, our method achieves near-upper-bound accuracy for surrogate label assignment while matching the performance of top post-hoc OOD detection methods. We demonstrate how \ourmethod scales in a real-world application, providing interpretable insights into model behavior under distribution shifts and helping identify potential biases and limitations in the dataset. This contributes to the robustness and trustworthiness of models, making it particularly valuable in safety-critical, time-sensitive, and low-data settings.
{
    \small
    \bibliographystyle{ieeenat_fullname}
    \bibliography{main}
}

\clearpage
\setcounter{page}{1}
\maketitlesupplementary

In this supplement, we first detail the datasets and models used, followed by a discussion of the introduced distribution shifts and their design rationale. Next, we evaluate the impact of various design choices, including layer selection, downsampling methods, classifiers, and surrogate label assignment hyperparameters. Finally, we present additional experimental results with detailed visualizations and performance metrics to provide deeper insights into the behavior and performance of our method.

\section{Datasets and Model Details}
\label{sec:supp_dataset}
\subsection{EuroSAT} 
\textit{EuroSAT}~\cite{eurosat} is a scene classification dataset derived from Sentinel-2 satellite images, covering various locations across Europe. It contains 27,000 images labeled into ten land-use and land-cover classes: Annual Crop, Forest, Herbaceous Vegetation, Highway, Industrial, Pasture, Permanent Crop, Residential, River, and Sea/Lake. The images have a spatial resolution of 10 meters.

We use a ResNet50 model pre-trained on ImageNet, modified to accept 13 input channels corresponding to Sentinel-2 spectral bands. The model is fine-tuned with a learning rate of 0.0001 and a batch size of 128. Training runs for up to 100 epochs with early stopping after 5 epochs of no improvement. Input images are normalized using channel-wise mean and standard deviation statistics.

\subsection{xBD}
\label{subsec:supp_dataset}
\textit{xBD}~\cite{xBD} is a semantic segmentation dataset for building damage assessment from satellite imagery. The dataset, collected from Maxar's Open Data Program, has images with a spatial resolution below 0.8 meters. It includes pre- and post-disaster images of hurricanes, floods, wildfires, and earthquakes, making it suitable for evaluating temporal and semantic shifts.

We simplify the damage assessment task into binary segmentation by reassigning damage levels: background (0) and levels 1-2 are grouped, while levels 3-4 form a high-damage class. This minimizes concept drift and ensures a fair evaluation of distribution shifts. We train a U-Net model with a ResNet50 backbone, pre-trained on ImageNet and configured for 3 input channels. Training uses a batch size of 32, a learning rate of 0.001, and runs for up to 50 epochs with early stopping after 5 epochs of no improvement. We reserve 10\% of the data for validation and normalize the input images by dividing pixel values by 255.

\subsection{FTW}
We follow the practices of the original study and use a U-Net model with an EfficientNet-B3 backbone for semantic segmentation on the FTW dataset. The model is configured with 8 input channels and outputs 3 classes: background, field, and field-boundary. We use class weights of [0.04, 0.08, 0.88] to address class imbalance. The learning rate is set to 0.001, and the loss function is cross-entropy. The number of filters is set to 64, and neither the backbone nor the decoder is frozen during training. We set the patience for early stopping to 100 epochs. The images are normalized by dividing pixel values by 3000.

For the classifier, we use logistic regression with a maximum number of iterations set to 500. We train the classifier with 500 ID samples and 1200 WILD samples. The number of clusters is set to 150, calculated as 0.3 times the total number of WILD samples. To reassign labels, we use an ID fraction threshold of 0.1, meaning that a cluster is assigned as OOD if ID samples comprise less than 10\% of the total samples in the cluster. The values of 0.3 and 0.1 are determined based on empirical observations gathered from extensive experiments on the xBD and EuroSAT datasets.

\begin{figure*}[h]
    \centering
    \includegraphics[width=1.0\linewidth]{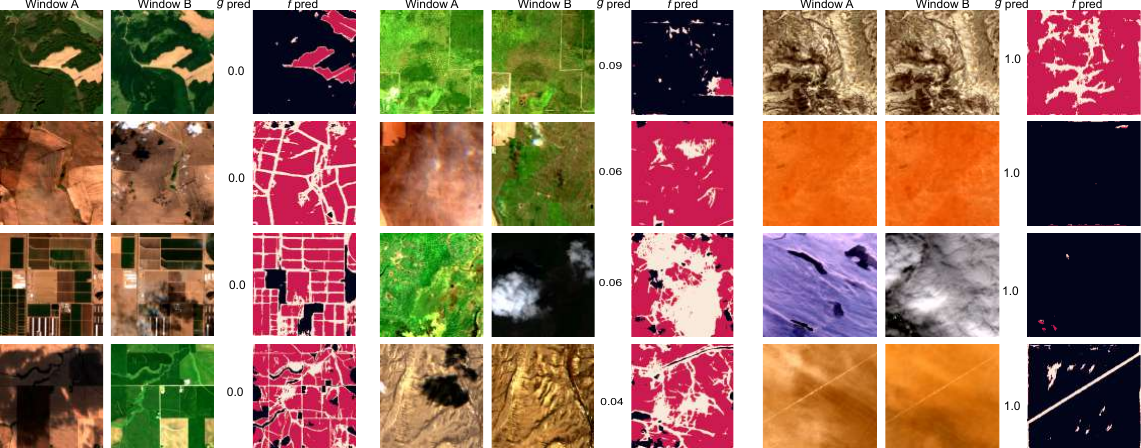}
    \caption{Deploying \ourmethod over FTW dataset: The input samples are from the collected WILD set, where the distribution is unknown. The figure shows Sentinel-2 images at two different times (planting season and harvesting season — Window A and Window B). When these windows are fed together into \( f \), the model outputs both the segmentation prediction and the OOD classifier \( g \)'s prediction score.}
    \label{supfig:tardis_pred_extra}
\end{figure*}

Figure~\ref{supfig:tardis_pred_extra} provides a visual illustration of the input samples from the WILD set, where the distribution is unknown. It displays the input Sentinel-2 image pair (Window A and Window B) alongside the OOD classifier \( g \)'s prediction scores and the DL model \( f \)'s predictions.

\subsection{Introducing Distribution Shifts to EuroSAT and xBD}
The combination of EuroSAT and xBD provides a diverse testbed for evaluating distribution shifts. EuroSAT represents regional imagery at medium spatial resolution, while xBD provides global imagery at very high resolution. Their differences in acquisition times, sensor parameters, processing levels, and the tasks they cover—land-cover classification (EuroSAT) and building detection (xBD)—make them complementary. Additionally, EuroSAT focuses on patch-level classification, while xBD involves pixel-level segmentation, enabling evaluations across different problem dimensions.

To evaluate our method, we introduce two types of distribution shifts: covariate and semantic (described in Table ~\ref{tab:covariate_shifts}). Our approach assumes that purposefully rearranging dataset splits creates measurable shifts between training and testing sets, driven by the logic of the split design.

\begin{figure*}[h]
    \centering
    \includegraphics[width=1\linewidth]{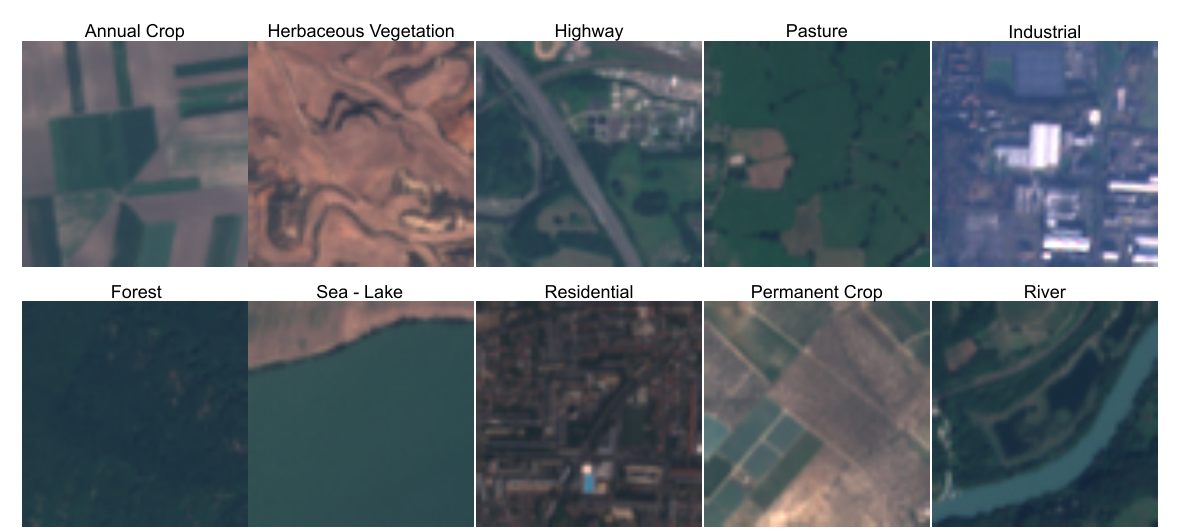}
    \caption{Examples from the EuroSAT dataset, with one sample from each class. These images highlight the spatial and semantic distinctions across classes.}
    \label{supfig:eurosat_dist_shift_viz}
\end{figure*}

\paragraph{EuroSAT Distribution Shifts.}
Figure ~\ref{supfig:eurosat_dist_shift_viz} shows one example from each of EuroSAT’s 10 classes, which differ spatially and semantically. For covariate shifts, we split the dataset by longitude at the midpoint of its spatial extent, using the western half for training and the eastern half for testing. This creates a shift based on spatial proximity.

For semantic shifts, we train the model on 9 classes and test it on the hold-out class, repeating this process for all classes. This ensures the model faces unseen scenarios during testing, providing a robust evaluation of its ability to handle semantic shifts.

\begin{figure*}[h]
    \centering
    \includegraphics[width=0.92\linewidth]{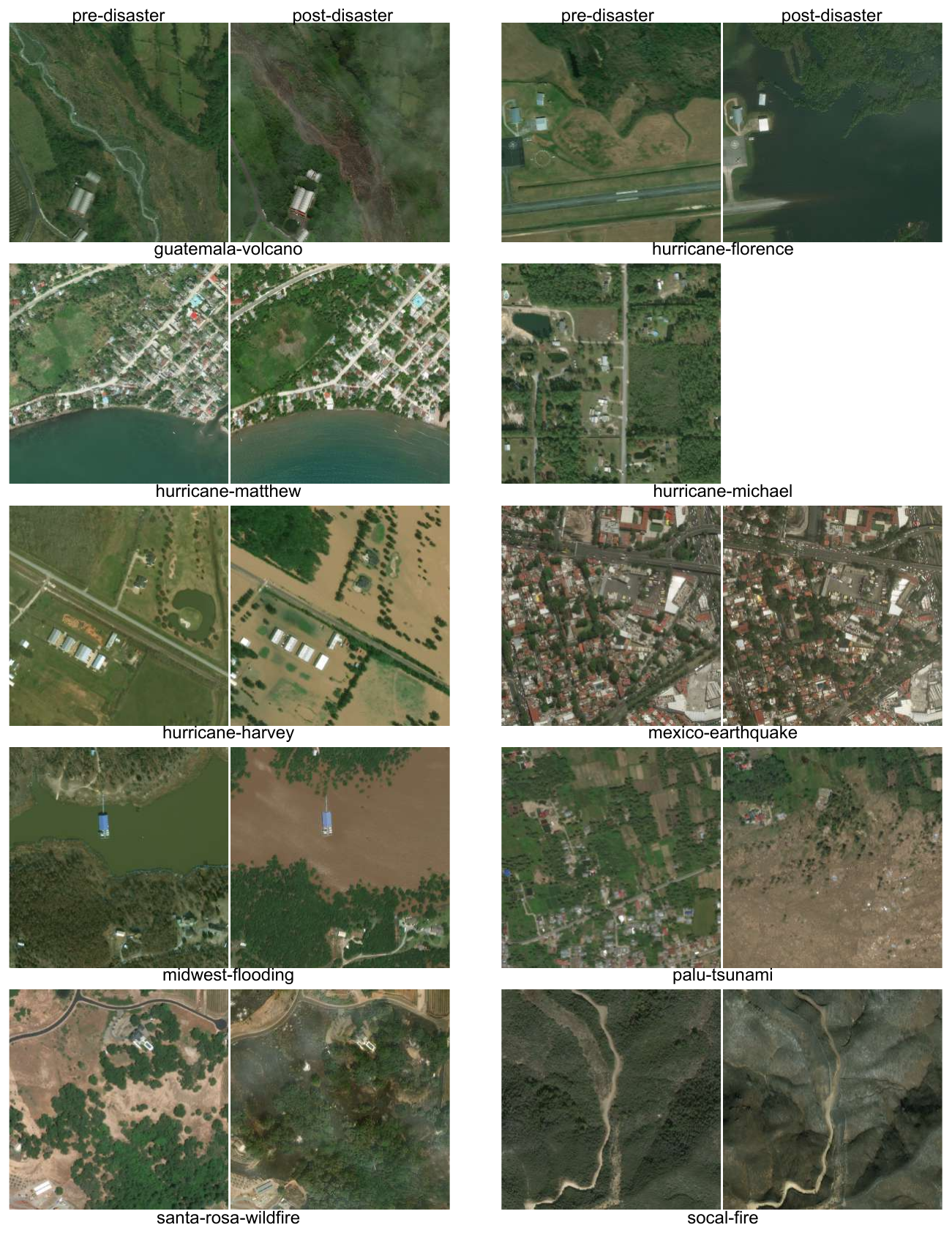}
    \caption{Examples from the xBD dataset, illustrating pre- and post-disaster images. These samples demonstrate the temporal and semantic differences between pre- and post-disaster scenes, highlighting the challenges posed by distribution shifts.}
    \label{supfig:xbd_dist_shift_viz_v2_png}
\end{figure*}

\paragraph{xBD Distribution Shifts.}
Figure ~\ref{supfig:xbd_dist_shift_viz_v2_png} illustrates the pre- and post-disaster image pairs in the xBD dataset. Temporal shifts arise from changes occurring between pre- and post-disaster images, while spatial and thematic shifts reflect differences in how disasters impact regions and leave varying degrees of visible marks. Using these inherent characteristics, we design covariate shift experiments for xBD.

\section{Design Choices}
\label{sec:supp_benchmark}
To better understand the impact of various design choices on the performance of our OOD detection method, we conduct a series of ablation studies. Specifically, we explore four key factors: (1) the choice of layer from which to extract feature representations (Section~\ref{sec:supp_layerbenck}), (2) the method used to downsample these feature maps (Section~\ref{sec:supp_downsamplingbench}), (3) the type of binary classifier \( g \) used to distinguish between surrogate-ID and surrogate-OOD samples (Section~\ref{sec:supp_gbench}), and (4) the selection of hyperparameters \textit{k} and \textit{T} for surrogate label assignment (Section~\ref{sec:supp_optuna_k_t}).

\subsection{Which Layer?}
\label{sec:supp_layerbenck}
Selecting the appropriate layer for activation extraction is crucial for accurate OOD detection. Prior works have emphasized the importance of this choice. For example, ASH achieves optimal performance on later layers like the penultimate layer, as earlier layers suffer from significant performance degradation during pruning~\cite{ash_djurisic23}. Similarly, ReAct performs best on the penultimate layer, where more distinctive patterns between ID and OOD data emerge~\cite{react_sun2021}. NAP-based OOD detection further highlights the variability in layer effectiveness, dynamically selecting top-performing layers based on validation accuracy~\cite{nap_ood}. Consistent with these findings, we observe that no single layer is universally optimal across all settings.

We benchmark FPR95 scores for OOD detection across the first convolutional layer, eight randomly selected intermediate layers, and the last convolutional layer. As shown in Table~\ref{suptab:layer_benchmark_eurosat} for the EuroSAT dataset and Table~\ref{suptab:layer_benchmark_xbd} for the xBD dataset, layer performance varies significantly. While late layers often perform well, early and middle layers frequently give competitive results, depending on the dataset and task. Based on these findings, we select the best-performing layer for each experiment.

For the large-scale FTW dataset, the lack of distribution shift information prevents evaluation of layer-specific performance for OOD detection. Therefore, based on the observation that many layer benchmarks perform optimally for middle layers, we select a middle convolutional layer, specifically \textit{`decoder.blocks.0.conv1`} from the U-Net model with an EfficientNet-B3 backbone.

\subsection{Which Downsampling Method?}
\label{sec:supp_downsamplingbench}
Having identified the layer to extract internal activations from, the next step is to look into the effect of downsampling these activations, which can reduce computational complexity and noise while retaining essential features for OOD detection. We explored four methods:

\begin{enumerate}
    \item \textbf{Mean and standard deviation (\textit{Mean Std})}: Computes the mean and standard deviation across the spatial dimensions (H, W) for each channel, providing two descriptive statistics per feature channel.
    \item \textbf{Average pooling (\textit{Avg Pool})}: Global average pooling was applied, reducing the activation to a single representative value per channel by averaging all spatial values.
    \item \textbf{Max pooling (\textit{Max Pool})}: Uses global max pooling to retain the maximum value from each spatial dimension, capturing the most prominent feature in each channel.
    \item \textbf{PCA-based reduction (\textit{PCA})}: Applies Principal Component Analysis to reshape the activation map into a vector and projects it into a lower-dimensional space with 10 components.
\end{enumerate}

We summarize the OOD detection performance across all experiments on the EuroSAT and xBD datasets under different downsampling methods in Table~\ref{suptab:downsampling_benchmark}, using the FPR95 metric. Max pooling consistently achieves the best performance across the majority of experiments, making it the preferred approach. We attribute its performance to its ability to retain the most prominent features in each channel, filtering out less significant information. This focus on salient patterns likely enhances the OOD classifier's capacity to distinguish between ID and OOD samples.

\subsection{Which Classifier?}
\label{sec:supp_gbench}
The next key design choice is the selection of the binary classifier \( g \), used to distinguish between surrogate-ID and surrogate-OOD samples based on their feature representations. The results, summarized in Table~\ref{suptab:g_benchmark}, report the mean performance across all experimental measurements along with the standard error of the mean to represent confidence intervals. We select Logistic Regression as it provides the best tradeoff between classification accuracy and prediction time. This balance is essential for scaling up the method, where both efficiency and accuracy are critical.

\subsection{Surrogate Label Assignment: Hyperparameter Search for \textit{k} and \textit{T}}
\label{sec:supp_optuna_k_t}

\ourmethod relies on a clustering-based approach in the activation space to assign surrogate ID and surrogate OOD labels. This process requires selecting two key parameters: the number of clusters (\textit{k}) to segment the activation space, and the ID fraction threshold (\textit{T}), which determines whether a cluster is assigned as surrogate ID or surrogate OOD. Clusters with an ID fraction above \textit{T} are assigned as surrogate ID, and those below \textit{T} are assigned as surrogate OOD.

The underlying assumption is that samples with similar distributions lie closer in the activation space than those from dissimilar distributions. Effectively clustering the activation space is critical, as the distributions of WILD samples are unknown during deployment, and it depends on the optimal choice of \( k \) and \( T \).

To develop insights into selecting \textit{k} and \textit{T}, we conduct controlled experiments on EuroSAT and xBD, where ID and OOD labels are known. In these experiments, we treat OOD labels as WILD and apply our clustering-based surrogate label assignment logic. By holding back the ground-truth WILD labels, we simulate real-world conditions while being able to evaluate the results against known labels.

The primary goal is to understand how to choose \textit{k} and \textit{T}, and whether there are patterns we can extrapolate to real-life deployment. For this, we first assign surrogate labels and calculate the ratio of \textit{k} to the total number of training samples and evaluate its effect on OOD detection performance (Accuracy, FPR95, and AUROC). We plot these metrics against the ratio of \textit{k}/total training samples, increasing \textit{k} until the ratio reaches 1. Theoretically, OOD detection improves with more clusters as this enables finer-grained clustering of the activation space, reducing the risk of including anomalies in ID clusters.

\begin{figure*}[htbp]
    \centering
    \includegraphics[width=1\linewidth]{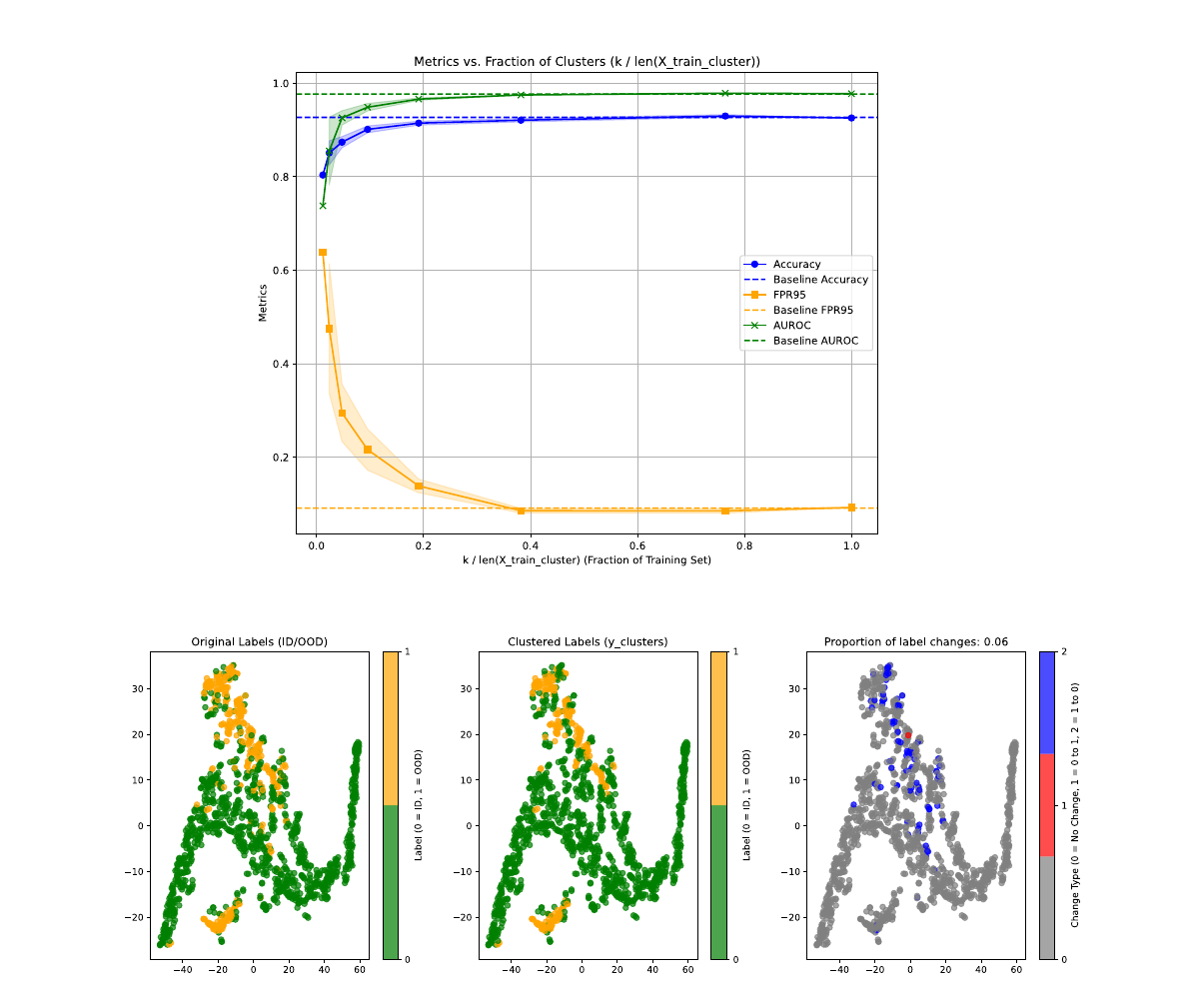}
    \caption{EuroSAT \textit{Pasture} experiment on surrogate label assignment. The upper plot shows the performance metrics (Accuracy, FPR95, AUROC) for the oracle classifier \( g_{\text{oracle}} \) and the surrogate classifier \( g^* \) as the ratio of clusters to training samples \( k / \text{len}(X_{\text{train}}) \) increases. As \( k \) grows, \( g^* \) gradually improves and approaches the performance of \( g_{\text{oracle}} \). The lower plot visualizes the feature space before and after clustering, showing how original ID and OOD labels are reassigned to surrogate ID and OOD labels based on the clustering logic.}
    \label{supfig:optuna_eurosat}
\end{figure*}

\begin{figure*}[htbp]
    \centering
    \includegraphics[width=1\linewidth]{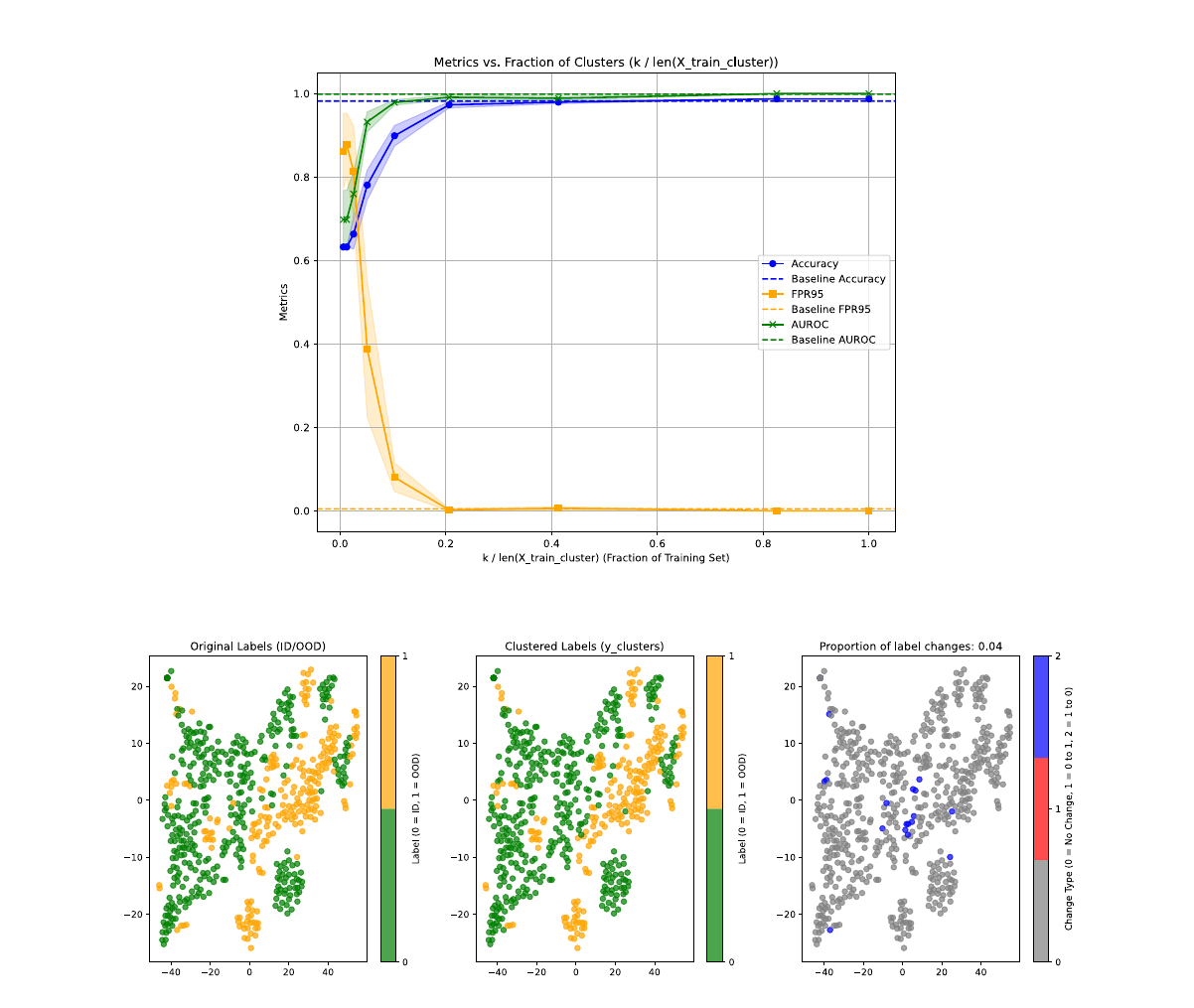}
    \caption{xBD \textit{Nepal Flooding-Midwest Flooding} disaster experiment on surrogate label assignment. The upper plot shows the performance metrics (Accuracy, FPR95, AUROC) for the oracle classifier \( g_{\text{oracle}} \) and the surrogate classifier \( g^* \) as the ratio of clusters to training samples \( k / \text{len}(X_{\text{train}}) \) increases. As \( k \) grows, \( g^* \) gradually improves and approaches the performance of \( g_{\text{oracle}} \). The lower plot visualizes the feature space before and after clustering, showing how original ID and OOD labels are reassigned to surrogate ID and OOD labels based on the clustering logic.}
    \label{supfig:optuna_xbd}
\end{figure*}

To establish a theoretical maximum (upper-bound performance), we also evaluate OOD classification with known ID and OOD labels, bypassing the need for clustering. This oracle performance is represented by horizontal dashed lines in Figure~\ref{supfig:optuna_eurosat} and Figure~\ref{supfig:optuna_xbd} (upper plots). The results for two representative experiments—one from EuroSAT and one from xBD—since all experiments show similar trends. We observe that the performance approaches the oracle boundaries when \textit{k} is approximately 0.3 times the total number of training samples. While performance improves as \textit{k} increases, a trade-off is required between performance and walltime as well as computational complexity. Based on this trade-off, we set \textit{k} to 0.3 for all experiments, including the large-scale deployment on FTW. Furthermore, we observe that our method is not highly sensitive to \textit{T}. As a result, we fix \textit{T} to 0.1 for all experiments, which is the value used in this initial investigation. We use the Optuna library to implement a Bayesian-based search algorithm. The composite objective function, which we minimize to determine the optimal number of clusters and ID fraction threshold, is detailed in Section~\ref{sec:method}.

Lastly, the gradual improvement in OOD detection performance with increasing \textit{k} supports our assumption that samples with similar distributions lie closer in the activation space than those with dissimilar distributions. The absence of degradation in performance further underscores the importance of activation-level clustering as a reliable proxy for domain estimation based on neighboring samples.

We set \textit{k} and \textit{T} as described and use t-SNE in the lower plots of Figure~\ref{supfig:optuna_eurosat} and Figure~\ref{supfig:optuna_xbd} to reduce the dimensionality of the activation spaces to 2D for visualization. When ID and OOD labels are known, the t-SNE plots show that only a small fraction of labels changes from the original labels. This demonstrates the effectiveness of the surrogate label assignment process described above.

\section{Further Experimental Results}
\label{supsec:addexpresults}
\begin{figure*}[t]
    \centering
    \includegraphics[width=1.0\linewidth]{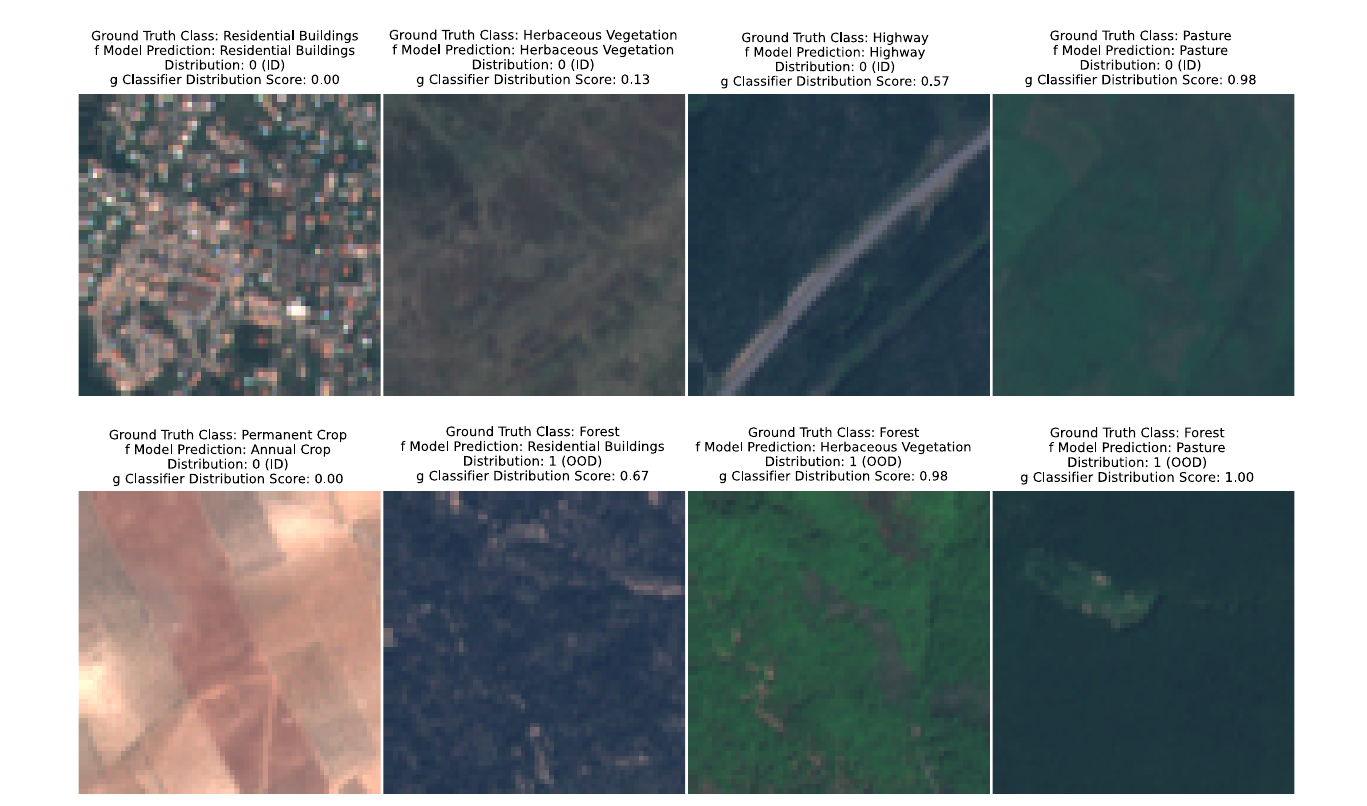}
    \caption{EuroSAT experiment with \textit{Forest} as the OOD class. The figure shows predictions of the DL model \( f \) and the OOD classifier \( g \), along with the ground truth class and distribution annotations. The first row represents samples where \( f \) makes correct class predictions, while the second row represents samples where \( f \) makes incorrect predictions. For each sample, we report both the ground truth distribution and the predicted distribution from \( g \).}
    \label{supfig:eurosat_shift_quantitative}
\end{figure*}

In Figure~\ref{supfig:eurosat_shift_quantitative}, we show the predictions of the DL model \( f \) and the OOD classifier \( g \), along with the ground truth class and distribution annotations for the EuroSAT experiment, where \textit{Forest} serves as the OOD class. The model \( f \) trains on 9 classes (excluding \textit{Forest}) and tests on \textit{Forest}. The first row shows correct predictions by \( f \), while the second row shows incorrect predictions. Even when \( f \) makes misclassifications, \( g \) accurately quantifies the distribution shifts in most cases. The performance of \( f \) on the test set is not directly measurable since the test uses a single unseen class. We report the performance of \( g \) as: Accuracy: \( 93.25\% \), ROC AUC: \( 98.86\% \), FPR95: \( 6.19\% \).

\begin{figure*}[ht]
    \centering
    \includegraphics[width=0.7\linewidth]{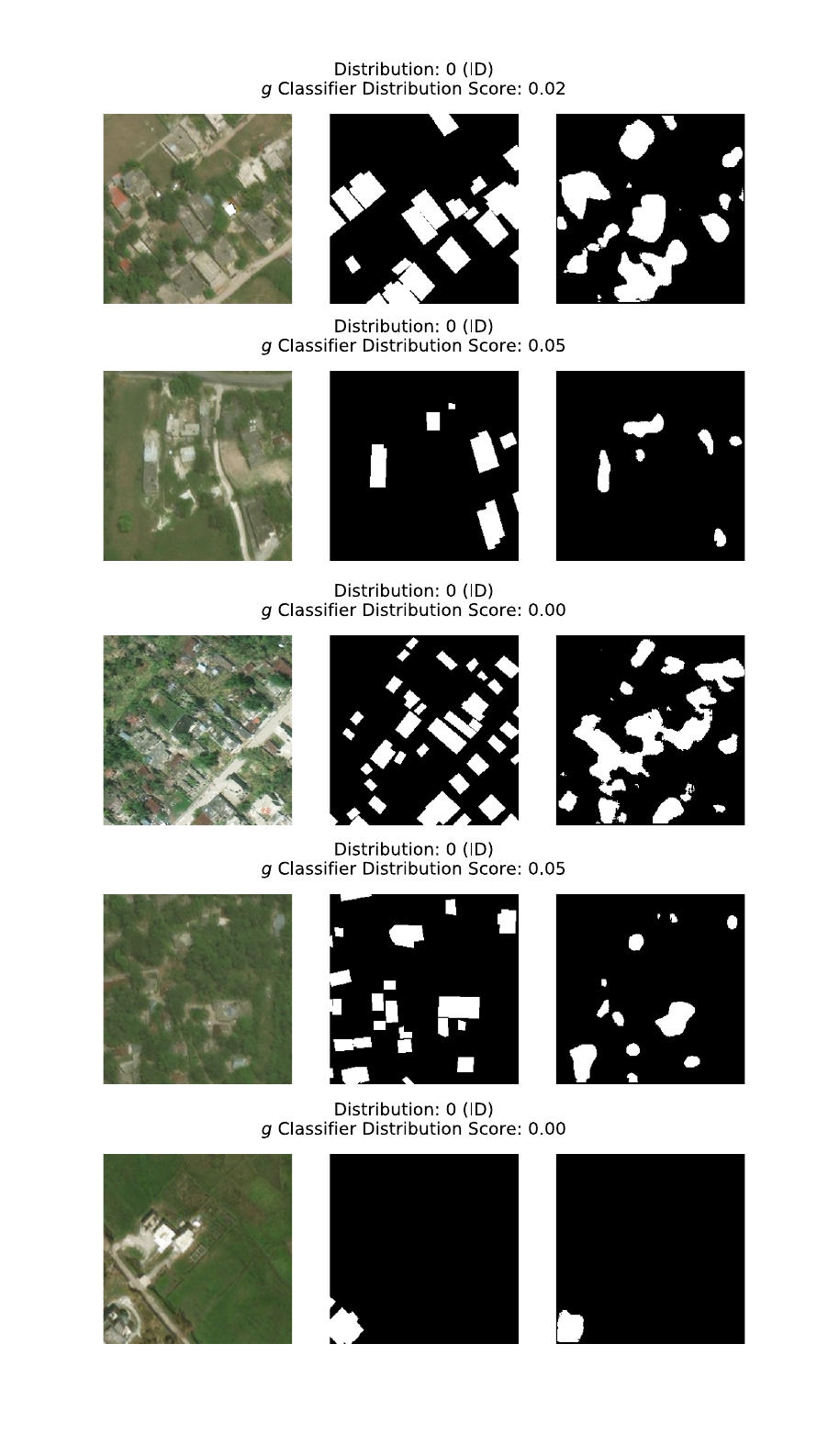}
    \caption{xBD experiment with \textit{Hurricane Matthew} as the ID samples. The figure shows the annotations and predictions of the DL model \( f \) and the OOD classifier \( g \). For each sample, we present \( f \)'s predicted class and \( g \)'s predicted distribution, along with the ground truth annotations.}
    \label{supfig:xbd_shift_quantitative_id}
\end{figure*}

\begin{figure*}[ht]
    \centering
    \includegraphics[width=0.7\linewidth]{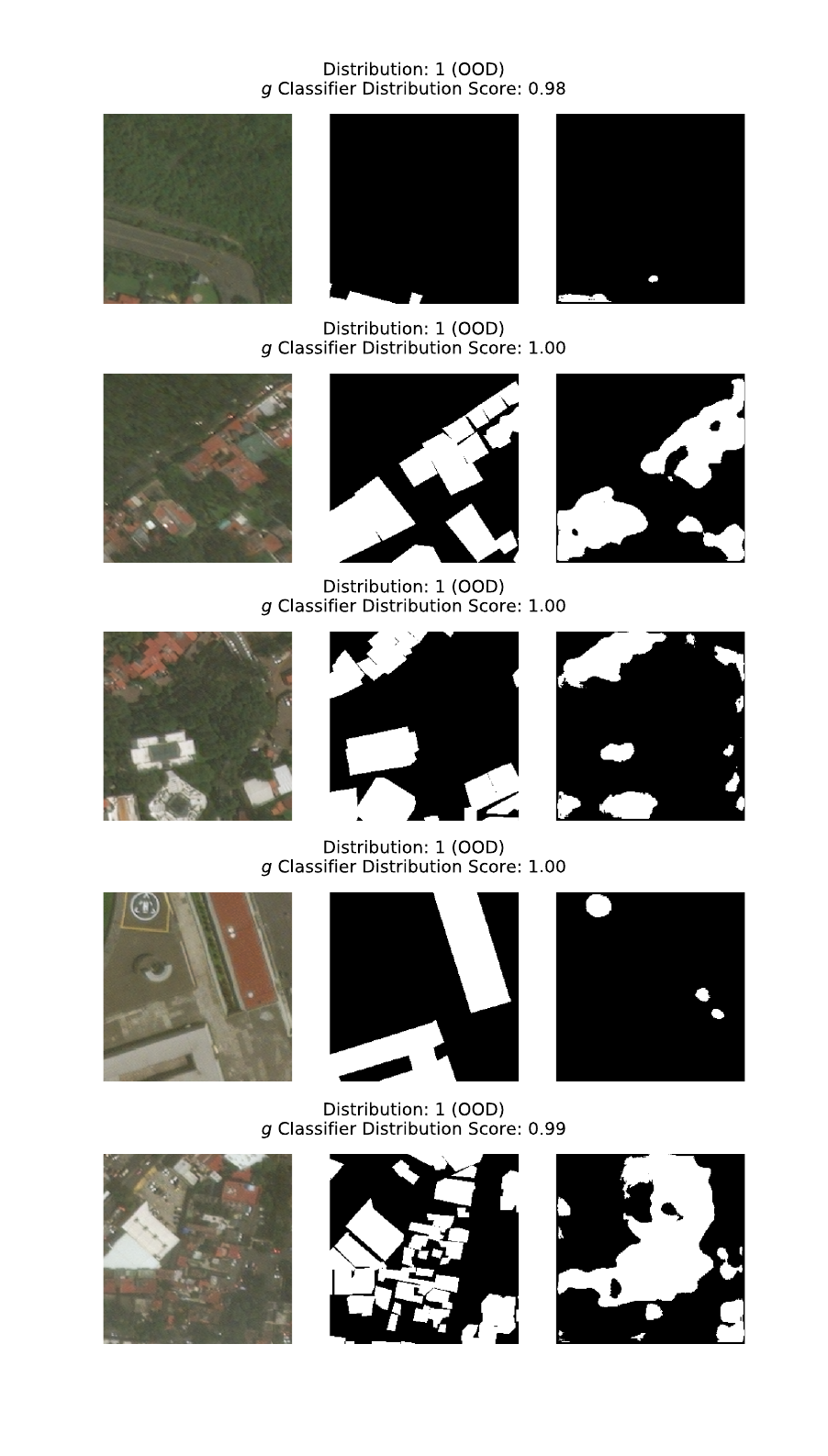}
    \caption{xBD experiment with \textit{Mexico Earthquake} as the OOD samples. The figure shows the annotations and predictions of the DL model \( f \) and the OOD classifier \( g \). For each sample, we present \( f \)'s predicted class and \( g \)'s predicted distribution, along with the ground truth annotations.}
    \label{supfig:xbd_shift_quantitative_ood}
\end{figure*}

For xBD, we present results where \( f \) is trained on \textit{Hurricane Matthew} (ID, Figure~\ref{supfig:xbd_shift_quantitative_id}) and tested on \textit{Mexico Earthquake} (OOD, Figure~\ref{supfig:xbd_shift_quantitative_ood}). Comparing the input images and masks between ID and OOD reveals that even when \( f \) performs suboptimally, \( g \) effectively quantifies the distribution shifts. The performance of \( f \) on the test set is as follows: Multi-class accuracy: \( 76.90\% \), Multi-class Jaccard index: \( 62.48\% \). We attribute \( f \)'s suboptimal prediction performance to the significant distribution shift between the training (\textit{Hurricane Matthew}) and testing (\textit{Mexico Earthquake}) datasets, and also to the fact that we reformulate the main task of damage classification to building detection (as described in Section \ref{subsec:supp_dataset}). The performance of \( g \) is: Accuracy: \( 98.06\% \), ROC AUC: \( 99.86\% \), FPR95: \( 0.00\% \).

\begin{table*}[htbp]
    \centering
    \begin{tabular}{>{\centering\arraybackslash}m{1.5cm} 
                    >{\centering\arraybackslash}m{1.2cm} 
                    >{\centering\arraybackslash}m{1.2cm} 
                    >{\centering\arraybackslash}m{1.2cm} 
                    >{\centering\arraybackslash}m{1.2cm} 
                    >{\centering\arraybackslash}m{1.2cm} 
                    >{\centering\arraybackslash}m{1.2cm} 
                    >{\centering\arraybackslash}m{1.2cm} 
                    >{\centering\arraybackslash}m{1.2cm} 
                    >{\centering\arraybackslash}m{1.2cm} 
                    >{\centering\arraybackslash}m{1.2cm}} 
    \toprule
    \textbf{Experiment} & \textbf{2/217} & \textbf{8/217} & \textbf{16/217} & \textbf{38/217} & \textbf{43/217} & \textbf{48/217} & \textbf{118/217} & \textbf{139/217} & \textbf{199/217} & \textbf{211/217} \\ \midrule
    Forest & 0.0625 & 0.01 & 0.00 & 0.00 & 0.00 & 0.00 & 0.0078 & 0.0156 & 0.0391 & 0.0391 \\
    HerbVeg & 0.2857 & 0.22 & 0.3095 & 0.22 & 0.2778 & 0.1429 & 0.07 & 0.1032 & 0.2460 & 0.2778 \\
    Highway & 0.8319 & 0.5462 & 0.6218 & 0.3529 & 0.3613 & 0.21 & 0.12 & 0.0840 & 0.1765 & 0.2437 \\
    Industrial & 0.2406 & 0.01 & 0.0376 & 0.0226 & 0.0226 & 0.00 & 0.0150 & 0.0226 & 0.0376 & 0.0075 \\
    Pasture & 0.1288 & 0.0909 & 0.12 & 0.1364 & 0.0985 & 0.03 & 0.0833 & 0.0227 & 0.1212 & 0.2273 \\
    PermCrop & 0.3554 & 0.2975 & 0.3140 & 0.2397 & 0.2314 & 0.14 & 0.12 & 0.1322 & 0.2066 & 0.1653 \\
    Residential & 0.2960 & 0.00 & 0.0160 & 0.0240 & 0.00 & 0.00 & 0.00 & 0.0160 & 0.0400 & 0.0480 \\
    River & 0.4688 & 0.07 & 0.2031 & 0.03 & 0.0938 & 0.0234 & 0.0078 & 0.0078 & 0.0625 & 0.0859 \\
    SeaLake & 0.00 & 0.00 & 0.00 & 0.00 & 0.00 & 0.00 & 0.00 & 0.00 & 0.00 & 0.00 \\
    AnnualCrop & 0.2879 & 0.00 & 0.0303 & 0.0606 & 0.0530 & 0.02 & 0.0379 & 0.0152 & 0.0682 & 0.0758 \\
    SpatialSplit  & 0.3182 & 0.20 & 0.4773 & 0.15 & 0.1970 & 0.0909 & 0.2197 & 0.2273 & 0.6364 & 0.7652 \\ \midrule
    \textbf{Avg ± Stdev} & 0.2978 ± 0.2223 & 0.1313 ± 0.1729 & 0.1936 ± 0.2133 & 0.1124 ± 0.1176 & 0.1214 ± 0.1264 & 0.0597 ± 0.0741 & 0.0620 ± 0.0696 & 0.0588 ± 0.0712 & 0.1486 ± 0.1801 & 0.1760 ± 0.2185 \\ 
    \bottomrule
    \end{tabular}
    \caption{FPR95 scores for OOD detection for experiments on EuroSAT dataset across the first convolutional layer, eight randomly selected layers, and the last convolutional layer. Notation in the header (e.g., X/Y) refers to the 'layer number / total number of layers.' The last row, labeled 'Avg ± Stdev,' provides the mean ± standard deviation of the scores for each layer across all experiments.}
\label{suptab:layer_benchmark_eurosat}
\end{table*}

\begin{table*}[htbp]
    \centering
    \begin{tabular}{@{}>{\raggedright\arraybackslash}p{3.1cm} 
                    >{\centering\arraybackslash}m{1.0cm} 
                    >{\centering\arraybackslash}m{1.0cm} 
                    >{\centering\arraybackslash}m{1.0cm} 
                    >{\centering\arraybackslash}m{1.0cm} 
                    >{\centering\arraybackslash}m{1.0cm} 
                    >{\centering\arraybackslash}m{1.0cm} 
                    >{\centering\arraybackslash}m{1.0cm} 
                    >{\centering\arraybackslash}m{1.0cm} 
                    >{\centering\arraybackslash}m{1.0cm} 
                    >{\centering\arraybackslash}m{1.0cm}@{}}
        \toprule
        \textbf{Experiment} & \textbf{3/223} & \textbf{60/223} & \textbf{112/223} & \textbf{146/223} & \textbf{148/223} & \textbf{162/223} & \textbf{176/223} & \textbf{187/223} & \textbf{208/223} & \textbf{216/223} \\
        \midrule
        Nepal Flooding \newline - Midwest Flooding & 1.0000 & 0.09 & 0.6986 & 0.7534 & 0.6986 & 0.7397 & 0.46 & 0.8767 & 0.8356 & 0.8493 \\
        Santa Rosa Wildfire \newline - Woolsey Fire & 0.5000 & 0.36 & 0.36 & 0.7222 & 0.6389 & 0.6389 & 0.6944 & 0.6111 & 0.58 & 0.9167 \\
        Hurricane Matthew \newline - Nepal Flooding & 0.3023 & 0.06 & 0.3488 & 0.4419 & 0.4884 & 0.32 & 0.1163 & 0.3488 & 0.2791 & 0.4419 \\
        Hurricane Matthew \newline - Mexico Earthquake & 0.11 & 0.1471 & 0.2353 & 0.6471 & 0.4118 & 0.50 & 0.3235 & 0.4706 & 0.6471 & 0.9706 \\
        Portugal Wildfire \newline (Pre-Post) & 0.38 & 0.9583 & 0.3889 & 0.8472 & 0.8750 & 0.9167 & 0.9722 & 0.77 & 0.9861 & 0.8472 \\
        \midrule
        \textbf{Mean ± Stdev} & 0.4585 ± 0.3343 & 0.3231 ± 0.3740 & 0.4063 ± 0.1735 & 0.6824 ± 0.1524 & 0.6225 ± 0.1818 & 0.6231 ± 0.2275 & 0.5133 ± 0.3316 & 0.6154 ± 0.2146 & 0.6656 ± 0.2686 & 0.8051 ± 0.2095 \\
        \bottomrule    \end{tabular}
    \caption{FPR95 scores for OOD detection experiments on the xBD dataset across the first convolutional layer, eight randomly selected layers, and the last convolutional layer. Notation in the header (e.g., X/Y) refers to the 'layer number / total number of layers.' The last row, labeled 'Avg ± Stdev,' provides the mean and standard deviation of the scores for each layer across all experiments.}
\label{suptab:layer_benchmark_xbd}
\end{table*}

\begin{table*}[htbp]
\centering
\begin{tabular}{@{}>{\raggedright\arraybackslash}p{3.5cm} 
                >{\centering\arraybackslash}m{1.5cm} 
                >{\centering\arraybackslash}m{1.5cm} 
                >{\centering\arraybackslash}m{1.5cm} 
                >{\centering\arraybackslash}m{1.5cm}@{}}
\toprule
\textbf{Experiment} & \textbf{Avg Pool} & \textbf{Mean Std} & \textbf{Max Pool} & \textbf{PCA} \\ 
\midrule
Forest & \textit{0.0859} & 0.4297 & \textbf{0.0234} & 0.8750 \\
HerbaceousVegetation & \textit{0.2937} & 0.8651 & \textbf{0.2698} & 0.9921 \\
Highway & 0.7899 & \textit{0.7311} & \textbf{0.7059} & 0.9412 \\
Industrial & \textit{0.1880} & 0.2857 & \textbf{0.0526} & 0.9925 \\
Pasture & \textbf{0.0909} & 0.6970 & \textit{0.2576} & 0.9924 \\
PermanentCrop & \textbf{0.3058} & 0.9008 & \textit{0.4215} & 0.9669 \\
Residential & \textit{0.2640} & 0.6640 & \textbf{0.2160} & 0.8480 \\
River & \textit{0.4922} & 0.6172 & \textbf{0.1563} & 0.9922 \\
SeaLake & \textbf{0.0000} & 0.0313 & \textbf{0.0000} & 0.9766 \\
AnnualCrop & \textit{0.2500} & 0.7576 & \textbf{0.0455} & 0.9924 \\
SpatialSplit & \textit{0.3182} & 0.5379 & \textbf{0.3030} & 0.9848 \\
Nepal Flooding - Midwest Flooding & \textbf{0.0000} & \textit{0.6575} & 0.9452 & 0.9726 \\
Hurricane Matthew - Nepal Flooding & \textbf{0.0233} & 0.9070 & \textit{0.5581} & 0.9535 \\
Hurricane Matthew - Mexico Earthquake & \textbf{0.0588} & 0.9118 & \textit{0.5882} & 1.0000 \\
Portugal Wildfire Pre - Portugal Wildfire Post & \textit{0.9028} & 0.9861 & \textbf{0.8472} & 1.0000 \\
\bottomrule
\end{tabular}
\caption{FPR95 scores for OOD detection across different downsampling methods. The table compares performances of average pooling, mean and standard deviation pooling, max pooling, and PCA for various experiments. Bold values indicate the best performance for each experiment, while italicized values represent the second-best performance.}
\label{suptab:downsampling_benchmark}
\end{table*}

\begin{table*}[htbp]
\centering
\begin{tabular}{@{}lcccc@{}}
\toprule
\textbf{Classifier} & \textbf{Accuracy$\uparrow$} & \textbf{ROC\_AUC$\uparrow$} & \textbf{FPR95$\downarrow$} & \textbf{Prediction Time (ms/sample)} \\ \midrule
KNeighbors & 92.23 ± 0.81 & 86.85 ± 1.07 & \textit{38.79 ± 2.02} & 73.00 ± 8.00 \\
GaussianNB & 84.28 ± 1.04 & 89.37 ± 0.91 & 32.89 ± 1.89 & 4.00 ± 1.00 \\
DecisionTree & 91.21 ± 0.93 & 77.43 ± 1.20 & 78.81 ± 2.41 & \textbf{2.00 ± 1.00} \\
ExtraTrees & \textbf{93.30 ± 0.67} & 91.29 ± 0.83 & \textit{28.53 ± 1.94} & 12.00 ± 3.00 \\
LogisticRegression & 87.67 ± 1.00 & \textit{93.33 ± 0.87} & 27.20 ± 1.98 & \textit{3.00 ± 1.00} \\
SVC & 91.54 ± 0.90 & \textbf{94.26 ± 0.72} & \textbf{19.87 ± 1.85} & 67.00 ± 12.00 \\
RandomForestUnbalanced & 92.54 ± 0.82 & 91.24 ± 0.80 & 30.98 ± 1.95 & 9.00 ± 2.00 \\
RandomForest & \textit{92.76 ± 0.75} & 91.11 ± 0.84 & 30.24 ± 1.91 & 8.00 ± 2.00 \\
AdaBoost & 92.85 ± 0.79 & 92.03 ± 0.82 & 29.84 ± 1.89 & 11.00 ± 3.00 \\
GradientBoosting & 92.92 ± 0.81 & 93.11 ± 0.85 & 30.47 ± 1.88 & 7.00 ± 2.00 \\ \bottomrule
\end{tabular}
\caption{Benchmark results of classifiers \( g \), including Accuracy, ROC AUC, FPR95, and prediction time. Values are reported as mean ± SEM over all experiments on EuroSAT and xBD. Bold indicates the best performance, and italics indicate the second-best performance. Prediction time is reported in milliseconds (ms/sample).}
\label{suptab:g_benchmark}
\end{table*}

\end{document}